%% file: egpaper_final.tex
\documentclass[10pt,twocolumn,letterpaper]{article}

\usepackage{iccv}
\usepackage{times}
\usepackage{epsfig}
\usepackage{graphicx}
\usepackage{amsmath}
\usepackage{amssymb}

\usepackage{framed,multirow}

\usepackage{booktabs}
\usepackage{subfigure}

\usepackage[breaklinks=true,bookmarks=false]{hyperref}

\iccvfinalcopy % *** Uncomment this line for the final submission

\def\iccvPaperID{****} % *** Enter the ICCV Paper ID here

\begin{document}

%%%%%%%%% TITLE
\title{Product Image Recognition with Guidance Learning and Noisy Supervision}

\author{Qing Li$^{1,2}$, Xiaojiang Peng$^2$, Liangliang Cao$^4$, Wenbin Du$^2$, Hao Xing$^3$, Yu Qiao$^2$\\
$^1$Southwest Jiaotong University, Chengdu, China\\
$^2$Shenzhen Institutes of Advanced Technology, CAS, China\\
$^3$Vipshop Inc., Guangzhou, China\\
 $^4$University of Massachusetts at Amherst, U.S.A.\\
%{\tt\small xj.peng@siat.ac.cn}
}

\maketitle

\begin{abstract}
This paper considers to recognize products from daily photos, which is
an important problem in real-world applications but also challenging due to background clutters, category diversities, noisy labels, etc.
We address this problem by two contributions. First, we introduce a novel large-scale product image dataset, termed as Product-90. Instead of collecting product images by labor-and time-intensive image capturing, we take advantage of the web and download images from the reviews of several e-commerce websites where the images are casually captured by consumers. Labels are assigned automatically by the categories of e-commerce websites. Totally the Product-90 consists of more than 140K images with 90 categories. Due to the fact that consumers may upload unrelated images, it is inevitable that our Product-90 introduces noisy labels.
As the second contribution, we develop a simple yet efficient \textit{guidance learning} (GL) method for training convolutional neural networks (CNNs) with noisy supervision.  The GL method first trains an initial teacher network with the full noisy dataset, and then trains a target/student network with both large-scale noisy set and small manually-verified clean set in a multi-task manner. Specifically, in the stage of student network training, the large-scale noisy data is supervised by its guidance knowledge which is the combination of its given noisy label and the soften label from the teacher network. We conduct extensive experiments on our Products-90 and public datasets, namely Food101,  Food-101N, and Clothing1M. Our guidance learning method achieves performance superior to state-of-the-art methods on these datasets.
\end{abstract}

%%%%%%%%% BODY TEXT
\input{intro2.tex}

\section{Related Work}
Our work is related to product image recognition and noisy data learning. In this section, we first review some related product image datasets and noisy datasets, and then present existing noisy data learning methods.

\subsection{Related Datasets}
\textbf{Product image datasets}. 
As for product images in computer vision and multimedia conmmunity, researchers mainly focus on the products of retails and groceries such as Supermarket~\cite{rocha2010automatic}, Grocery Products~\cite{george2014recognizing}, Gorzi-120~\cite{merler2007recognizing}, Feribur Groceries~\cite{jund2016freiburg}, RPC~\cite{wei2019rpc}.
\textbf{Supermarket}~\cite{rocha2010automatic} is introduced for automatic fruit and vegetable classification from images. It has 15 product categories with 2,633 images captured under diverse conditions. %The task, however, has nearly been solved as implemented solution achieves classification error under 2% on the dataset.
\textbf{Grocery Products}~\cite{george2014recognizing} is another dataset aiming at grocery product recognition. It contains 80 grocery product categories with 8,350 training images and 680 test images. %The training images are downloaded from the web, and the test images are collected in natural shelf scenario. %Normally the products are neatly placed on the shelf, which is different from the realistic checkout scenario where the products are freely placed on counter in clutter.
\textbf{Grozi-120}~\cite{merler2007recognizing} is a dataset proposed for groceries recognition in natural environment. It contains 120 grocery product categories. For each product category, two types of images are collected, one from the web, the other from inside a grocery store. In total, 11,870 images are collected with 676 from the web and 11,194 from the store. %Traditional al- gorithms, such as color histogram matching, SIFT matching, and boosted Haar-like features, are applied to the dataset for performance evaluation.
\textbf{Freiburg Groceries}~\cite{jund2016freiburg} is another grocery dataset comprising 5,021 images of 25 grocery classes. The images are divided into two sets: a training set that consists of 4,947 images taken by smartphone cameras, each containing
one or more instances of one class; a test set with 74 images of 37 clutter scenes, each containing objects of multiple classes.
\textbf{RPC}~\cite{wei2019rpc} is a recently-published retail product image dataset aiming at automatic checkout application. This dataset also provides images of two different types. One type is taken in a controlled environment and only contains a single product.  Another type represents images of user-purchased products and these images usually include multiple products. In total, it contains 83,739 images of 200 fine-grain classes. %These fine-grain classes are belong to 17 meta categories, namely puffed food, dried fruit, dried food, instant drink, instant noodles, dessert, drink, alcohol, milk, canned food, chocolate, gum, candy, seasoner, personal hygiene, tissue, stationery.  
%Perhaps the most related dataset is the \textbf{ALISC} dataset\footnote{https://tianchi.aliyun.com/competition/entrance/231510/information}, which contains 2 million product images with 10 general product categories (like shirt dress) and 676 subcategories~\cite{kyaw2017matryoshka}. However, their images are mostly collected from the online e-shops where images are well-captured..  
Different from these retail or grocery product image datasets, our proposed Product-90 is a \textit{label noisy} dataset collected by mobile cameras in daily life, and contains categories from retail products to clothing and shoes.  Liu \textit{et al.}~\cite{liu2012street} also propose a daily
photo dataset but it limited on clothing images.

\textbf{Noisy datasets}.
In recent research, both synthetic and real-world noisy datasets are widely used. For example, MNIST and CIFAR-10 are used as synthetic noisy datasets in \cite{patrini2017making,sukhbaatar2014training}. Synthetic label noise usually mimics random class noise and confusing class noise by corrupting the original clean datasets. To explore noisy data learning methods, three real-world noisy datasets are introduced more recently. Xiao \textit{et al.}~\cite{xiao2015learning} present the Clothing1M fashion image dataset which consists of 14 classes with more than a million images crawled from online shopping websites. Li \textit{et al.}~\cite{li2017webvision} introduce the WebVision dataset which contains 2.4M noisy labeled images crawled from Flickr and Google using the ILSVRC taxonomy~\cite{deng2009imagenet}. Lee \textit{et al.}~\cite{lee2017cleannet} collect the Food-101N dataset which contains 310k images from Google, Bing, Yelp, and TripAdvisor using the Food-101 taxonomy~\cite{bossard2014food}.
Our Products-90 is related to Clothing1M but contains much more categories including clothing, bags, jewelry, shoes, home products, personal care products, stationery, etc. Meanwhile, the Products-90 is crawled from the customer reviews of online shopping websites which includes more complex background clutter and noise.

\subsection{Noisy Data Learning Methods}
%Some works focus on how to filter noise in constructing datasets from the web~\cite{yao2017exploiting}, while recently learning with noisy data directly has been vastly studied on the literature of machine learning and computer vision. Nettleton \textit{et al.}~\cite{nettleton2010study} investigate the impact of attribute noise (e.g., Gaussian noise in data points) and class noise (i.e., label noise) on four popular supervised learners.  
We focus on the label noise problem and refer to \cite{frenay2014classification} for a comprehensive overview. 
Methods on learning with label noise can be roughly grouped into three categories: noise-robust methods, semi-supervised noisy data learning methods, and noise-cleaning methods.

\textbf{Noise-robust methods}. The noise-robust or noise-tolerance learning methods are assumed to be not too sensitive to the presence of label noise, which directly learn models from the noisy labeled data~\cite{joulin2016learning,krause2016unreasonable,misra2016seeing,patrini2017making,lu2017nuclear}.
Nettleton \textit{et al.}~\cite{nettleton2010study} show that the Naive Bayes probabilistic learner is less sensitive to label noise. Manwani~\cite{manwani2013noise} present a noise-tolerance algorithm under the assumption that the corrupted probability of an example is a function of the feature vector of the example. Mnih \textit{et al.}~\cite{mnih2012learning} propose two robust loss functions  to deal with label noise.
With synthetic noisy labeled data, Rolnick \textit{et al.}~\cite{rolnick2017deep} demonstrate that deep learning is robust to noise when training data is sufficiently large with large batch size and proper learning rate. Guo \textit{et al.}~\cite{guo2018curriculumnet} develop a curriculum training scheme to learn noisy data from easy to hard. %cleanNet mentorNet adjust the loss weights of mislabeled samples
Jiang \textit{et al.}~\cite{jiang2017mentornet} design a MentorNet to adjust the loss weights of noisy samples in the training process.

% In the previous research works, the deep CNNs framework is not extensively used for computer vision tasks. Directly learning discriminative methods from noisy labeled data has been proposed. Many of the experiments demonstrate that these methods are still affected by label noise and overfit with the noisy labels in the training process \cite{nettleton2010study}.  To better handle label noise and enhance the classification accuracy, several works depend on training classifiers with the noise-robust algorithms \cite{beigman2009learning,teng2000evaluating}. The noise-robust algorithms is not really deal with noise and is are still affected by label noise, which is illustrated in \cite{teng2001comparison}. After that, the label noise-tolerant methods is proposed to make use of some side information like the noisy rate in each lass to design models that account for the label noise \cite{manwani2013noise}. Another solution is used the label cleansing algorithms to remove or correct mislabeled data \cite{ brodley2011identifying}.   Following the review \cite{nettleton2010study}, it shows that these methods seem to be adequate only for simple cases of label noise, which can be safely managed by overfitting avoidance. Therefore, some label cleansing methods are proposed to remove or amend the noisy labels \cite{brodley2011identifying}.

\textbf{Semi-supervised noisy data learning methods}.
Semi-supervised methods aim to improve performance using a small manually-verified clean set. These methods usually obtain higher performance than the other methods since extra human supervision is added. 
Lee \textit{et al.}~\cite{lee2017cleannet} train an auxiliary CleanNet using manually-verified data to detect label noise and adjust the final sample loss weights. 
Similarly, Veit \textit{et al.}~\cite{veit2017learning} also use the clean set to train a label cleaning network but with a different architecture. These methods assume there exists such a label mapping from noisy labels to clean labels. Xiao \textit{et al.}~\cite{xiao2015learning} mix the clean set and noisy set, and train an extra label noise type CNN and a classification CNN to estimate the posterior distribution of the true label. \textit{Our guidance learning belongs to the semi-supervised noisy data learning which leverages a teacher-student training strategy to take full use of the whole data space (noisy set and clean set) and the student network trades off the noisy ground truths and soften labels by guidance knowledge.}

\textbf{Noise-cleaning methods}. Noise-cleaning methods aim to identify and remove or relabel noisy samples with filter approaches~\cite{miranda2009use}.
Brodley \textit{et al.}~\cite{brodley1999identifying} propose to filter noisy samples using ensemble classifiers with majority and consensus voting. Sukhbaatar \textit{et al.}~\cite{sukhbaatar2014training} introduce an extra noise layer into a standard CNN which adapts the network outputs to match the noisy label distribution. Daiki \textit{et al.}~\cite{tanaka2018joint} propose a joint optimization framework to train deep CNNs with label noise, which updates the network parameters and labels alternatively.
Based on the consistency of the noisy groundtruth and the current prediction of the model, Reed \textit{et al.}~\cite{reed2014training} present a `Soft' and a `Hard' bootstrapping approach to relabel noisy data. Similarly, Li \textit{et al.}~\cite{li2017learning} relabel noisy data using the noisy groundtruth and the current prediction adjusted by a knowledge graph constructed from DBpedia-Wikipedia. Our guidance learning framework is also related to \cite{reed2014training} and \cite{li2017learning} but differs in that \textit{i) we consider a small clean set instead of noisy data only which inherits the advantage of semi-supervised methods and ii) we train the teacher model with the full dataset and the student model with guidance knowledge in a multi-task learning manner.}

\begin{figure*}[t]
\centering
  \includegraphics[width=0.9\textwidth]{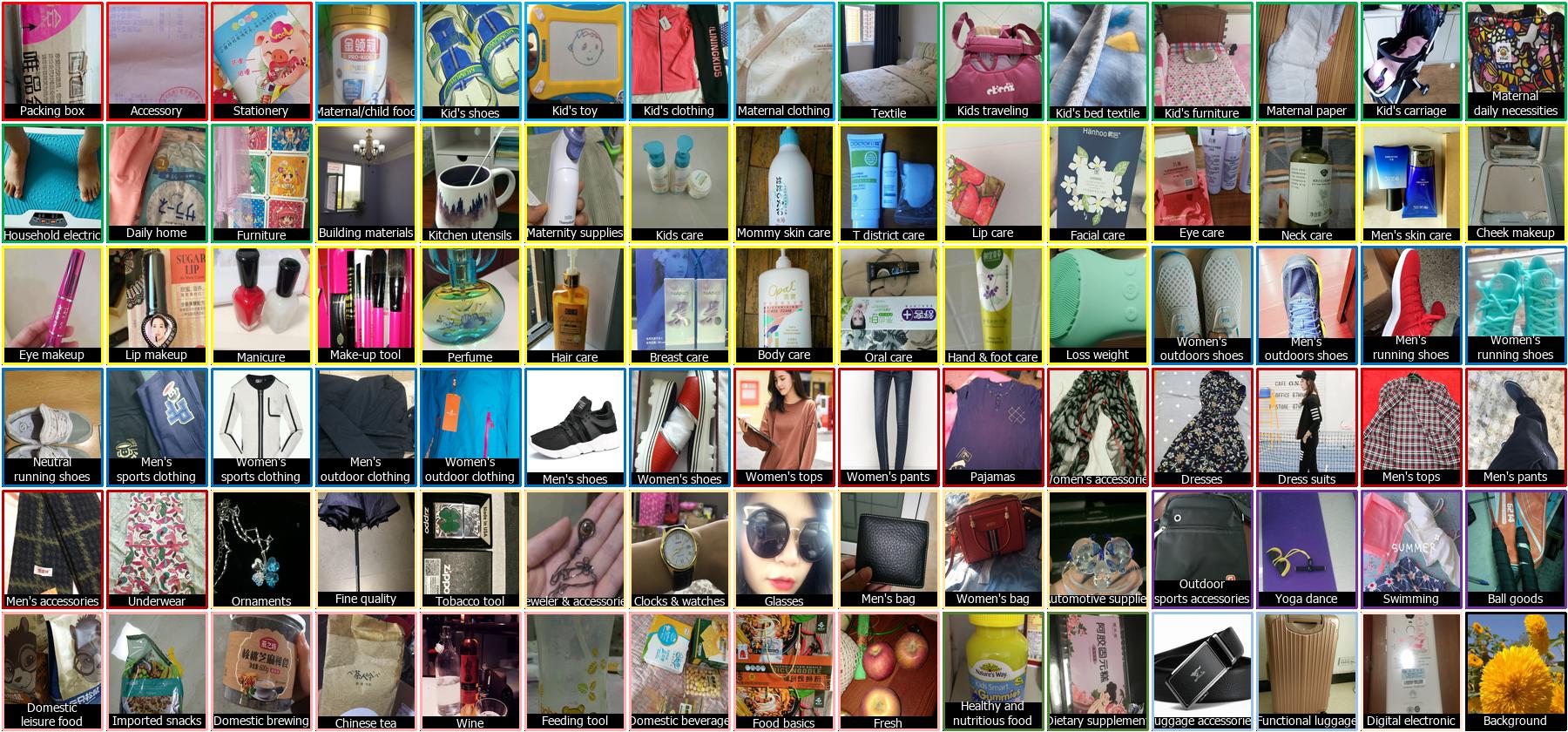}
  \caption{Illustration of the Products-90 dataset. Each image represents one class which is selected from clean data. Different image boundary colors correspond to different meta categories. (Zoom in for better view.)}\label{fig:figure3}
\end{figure*}

\begin{figure*}[t]
\centering
  \includegraphics[width=1\textwidth]{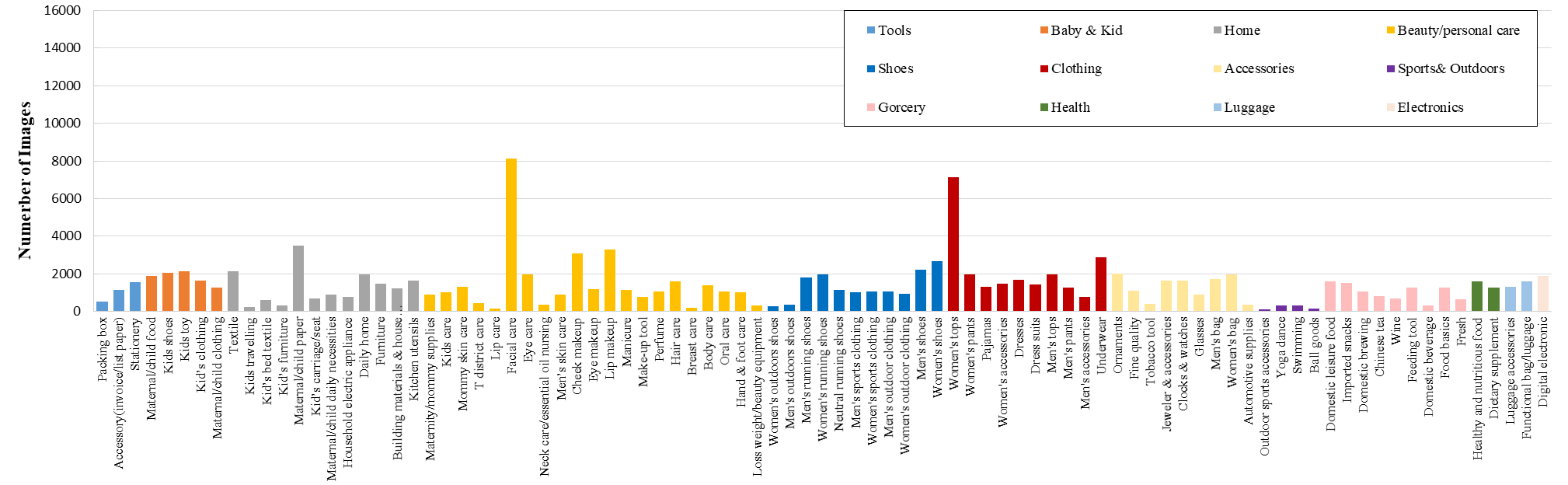}
  \caption{Statistics of the collected Products-90. Each color indicates a meta category. (Zoom in for better view.)
  }
  \label{fig:figure6}
\end{figure*}

\section{The Product-90 Dataset}
%\llc{I would name section III as "problem" instead of "proposed dataset. The whole section should be adjusted: This paper is not reporting performance on an existing dataset. Instead, this paper is solving a problem which has been ignored by academia. For this goal, we build a dataset, which reflect the challenges for product image recognition.}
%We address the problem of recognizing products from consumer photos with noisy supervision. 
%In this section, we first detail our collected Product-90, and then compare it to other related datasets.

Considering the applications of generic product image recognition, we collect the Products-90 dataset. It is collected by crawling images in consumer reviews from several e-shopping websites. Labels are assigned by categories provided by the sellers like in \cite{xiao2015learning}. Several image samples along with the annotation are shown in Figure \ref{fig:figure3}. The 90 classes are mainly selected according to the categories of these websites and filtered by their definition ambiguity. We keep the original fine-grain classes since it is useful in practice. These product classes can be mainly grouped into 13 meta categories, namely tools, baby and kids, home, beauty and personal care, shoes, clothing, accessories, sports/outdoors, grocery, health, luggage, electronics, and background. There exist fine-grain classes in most of the meta categories. For example, there are 21 categories in the beauty and personal care including eye makeup, lip makeup, cheek makeup, facial care, eye care, lip care, etc.

Our current version of Products-90 consists of 142,466 images, with hundreds or thousands of samples for each class. 
Figure \ref{fig:figure6} shows the statistics of Products-90 where each color corresponds to a certain meta categories. 
All categories are relatively balanced except for the `facial care' and `women tops'. To separate product classes from non-product class, the background category is added by crawling scene images (keywords such as landscape, building) in several searching engines. The number of images for the background is comparable to the one of the largest class (i.e. facial care).
%Our current version of Products-90 consists of 142,466 images, with hundreds or thousands of samples for each class.

\textbf{Protocols}.  Due to the fine-grain classes in Products-90, we find that it is hard to relabel or refine manually as in \cite{xiao2015learning}. Instead, we manually verify about 17K of all the images in the dataset which are correctly-labeled. We further split it into training ($D_c$) and test sets, which contains 8,795 and 8,787 images, respectively. This small training subset is used as clean set. The remaining of Products-90 is used as noisy training set ($D_n$). We report the overall accuracy on the clean test set.

\begin{figure*}[tp]
%\centering
\centering{
  \includegraphics[width=0.8\linewidth]{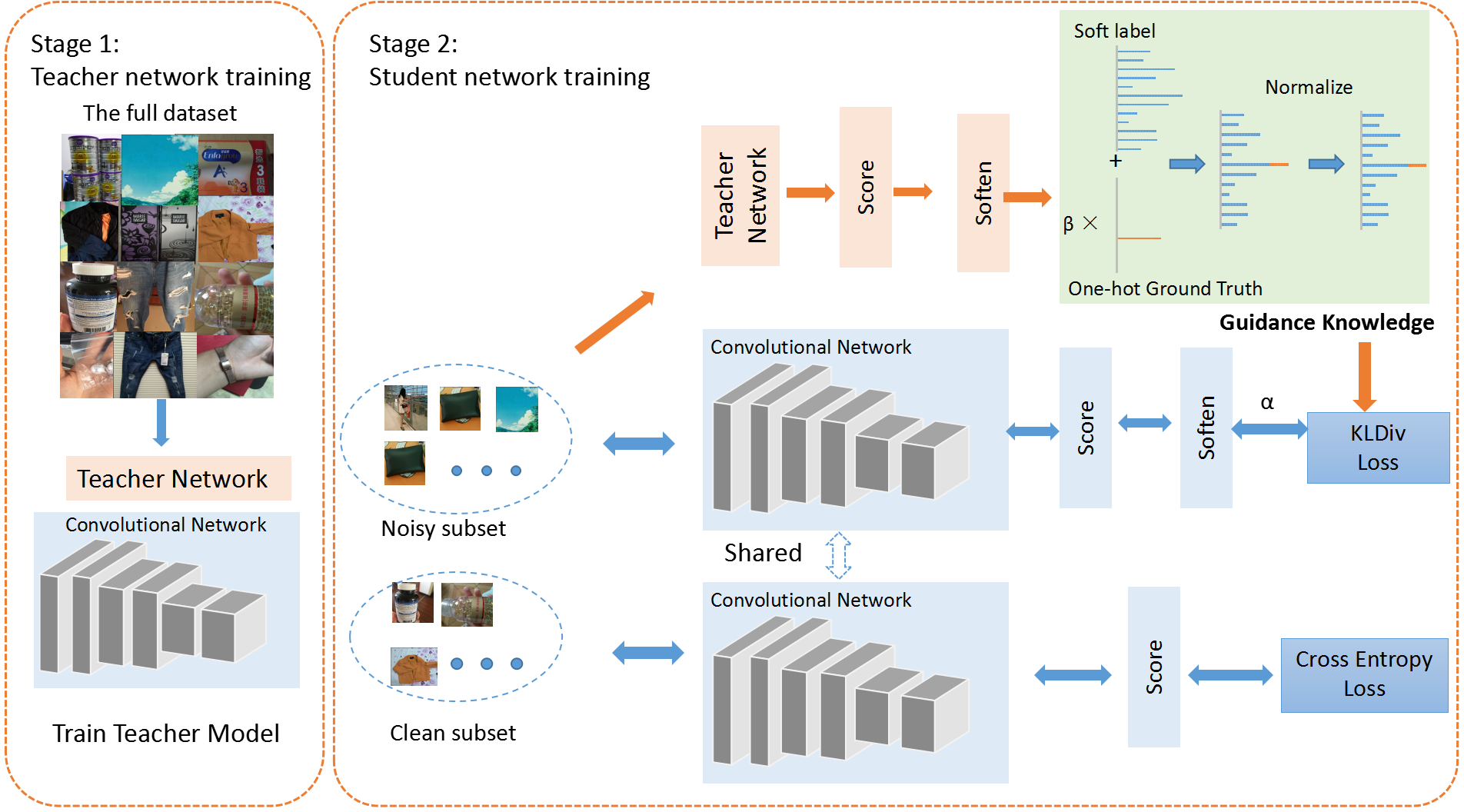}\\
  \caption{The proposed guidance learning framework. At the first stage, we utilize all training data to train a teacher model. At the second stage, we separate the training data into a noisy subset and a clean subset to train the student network with a multi-task learning mechanism.}\label{fig:figure2}
  }
\end{figure*}
\section{Guidance Learning}
We propose the guidance learning framework to deal with the problem of product image recognition with noisy supervision. The guidance learning framework is illustrated in Figure \ref{fig:figure2}.
Our framework consists of two stages: 1) teacher network training 2) student network training.
 In the first stage, we use all the training data to train a basic CNN model, which is called the teacher model.
 In the second stage, we use the noisy training dataset and the clean training dataset to train a student model in a multi-task learning manner.
%  The teacher model is brought into transfer the guidance knowledge to supervise the training of the student network. During learning the process of guidance knowledge, we can utilize the soften target and the modified ground truth label to generate a guidance soft signal. Then, we employ the guidance soft signal to guide the student model to distill the noisy label and boost the clean label. The student is monitored in the training progress and determined to learn the guidance soft signal information by the teacher. In addition, we use the guidance soft signal to compute the soften signal loss and joint with the cross-entropy.

\subsection{Teacher Network}
The teacher network is trained on the full dataset which contains mislabeled and correctly-labeled samples.
%The teacher network is to implement a basic CNNs to train visually grounded image classifier for a set of visual concepts. Therefore, it is considered to be an $L$ way multi-class classification problem.
More specifically,
%Our purpose is to built an teacher classifier network by using the deep CNNs model
given dataset $D=\{( x_{i},y_{i}) \mid i = 1 \ldots N \}$, where $ x_{i}$ denotes the $i-$th observed image and the corresponding label $y_{i} \in \left\{ 1,\ldots,C \right\}$ and $C$ is the number of the categories, we use all the training data $D$ to learn the teacher model.

%which is then applied to be a teacher model to guiding the student network training. The teacher model is used to compute a soften target from the deep representation of each image in all selected training set.

%About the function $\mathcal{L}$, the empirical risk of the classifier $f$ is defined as $R_{\mathcal{L}}(f) = \mathbb{E_{D}}[\mathcal{L}(f(x),y_{x})]$, where the expectation is over the empirical distribution.
% To compute the loss function for classification task, we use the most common cross entropy in this case, as follow:
 At this stage, the teacher network training is considered identical with the classical classification problem, assuming all the samples are correctly labeled.
 The loss function of teacher network is the cross entropy between the softmax output ${p}$ and the ground-truth distribution over labels ${q}$.
\begin{equation}\label{eq:CE}
%R_{\mathcal{L}}(f) = \mathbb{E_{D}}[\mathcal{L}(f(x;\theta),y_{x})] = -\frac{1}{N}\sum_{i=1}^{N}\sum_{j=1}^{L}y_{ij}\log f_{j}(x_{i};\theta)
%\mathcal{L}_{teacher} = -\frac{1}{N}\sum_{i=1}^{N}\sum_{j=1}^{C}\mathbf{y}_{ij}\log \mathbf{p}_{ij}
\mathcal{L}_{teacher} = -\sum_{i=1}^{C}{q}_{i}\log{p}_{i}
\end{equation}
where ${q}_{i}$ is the ground-truth distribution of the $i$th true class label. %, $y_{j} \in [0,1]^{L}$ such that $1^{T}y_{i} = 1 \forall i$,
%  and $p_{j}$ denotes the $j'th$ element of p.
%
%  Note that, $\sum_{i=1}^{N}f_{j}(x_{i};\theta) = 1$, and $f_{j}(x_{i};\theta) \geq 0, \forall j,i,\theta$.

\subsection{Student Network}
Once we finished the first stage, we obtain a teacher model which contains implicit information of the dataset.
%Two strategies are employed to tackle this dilemma.
%First, we
At the second stage, we train a target network in a multi-task manner with the large-scale noisy set and a small clean training set. We refer this set as the clean subset and the remaining noisy dataset as the noisy subset. We denote the clean training subset  $D_{c}=\left\{ \left ( \hat{x}_{i},y_{i} \right ) \mid i = 1 \ldots K \right\}$ and noisy subset
$D_{n}=\left\{ \left ( x^{\ast}_{i},y^{\ast}_{i} \right ) \mid i = 1 \ldots M \right\}$.
In our case, the clean training data is a small portion of the training data.
We have $N = \left|D_{c} \right| + \left|D_{n} \right|$ with $\left|D_{c} \right| \ll \left|D_{n} \right|$.
Since the clean subset is built based on the selected samples with the right annotation, the noisy subset and the clean subset could have different distribution. However, the clean training subset share the same distribution with the clean test set.
To take full advantage of the whole training dataset, we propose the guidance learning method which leverages different supervision information for the noisy subset and clean subset, and combine them with a multi-task learning (MTL) framework. %In the case of MTL, the auxiliary tasks cause the model to prefer hypotheses that explain more than one task which leads to solutions that generalize better~\cite{DBLP:journals/corr/Ruder17a}.
%\llc{Do we really need mention multi-task learning? The connection is not very obvious to me. If we must do, we should have cited multi-task learning paper}
%The way of building the clean subset is not necessarily the same as in our setting. The noisy subset could even come from different datasets or have different number of categories. Either case could be dealt with the proposed multi-task learning framework.

%Our new method is named \textsl{Guidance Learning}.
For the clean subset, it is natural to choose the traditional cross entropy loss for
training since we are confident about the sample labels. 
For the noisy subset, instead of using the original label which may be totally wrong, we 
%propose a novel guidance knowledge for this case. Intuitively, the guidance knowledge is designed to
aim to alleviate the effects of the noisy label and to maintain the right label functioning well at the same time.
To achieve this goal, we resort to the noisy labels as well as the knowledge included in the soften predictions of the teacher network.

Specifically, we first input the noisy subset into the teacher network to obtain logit scores $z_{i}\in R^C$ for the \textit{i}-th sample. Then, we feed the predictions into the softmax layer with a score-soften operation to obtain the target soft probabilities ${p}_{i}$. In the soften operation, inspired by \cite{hinton2015distilling} we introduce a temperature to transfer the knowledge of teacher. The target soft probability $p_{i}$ is defined as follows,

\begin{equation}\label{eq:equ1}
p_{i} = \frac{exp(z_{i}/T)}{\sum_{j}exp(z_{j}/T)},
\end{equation}
where ${z}_{i}$ is the pre-softmax activations of teacher network and T is the temperature which softens the signals. As mentioned in \cite{hinton2015distilling}, the soften operation can provide more information or knowledge about the model's prediction.

To achieve the final soft target of a sample, we fuse the knowledge from the teacher network  ${p}_{i}$ and its noisy label $ {y}_{i}$ %\llc{Is $t_i$ the same as $y_i$?} 
which is a one-hot vector as follows,
%\llc{why one-hot vector?} 
\begin{equation}\label{eq:equ2}
{g}_{i} = \dfrac{1}{(1+\beta)}({p}_{i} + \beta {y}_{i}),
\end{equation}
where $\beta$ is a trade-off weight of the two parts. We call ${g}_{i}$ as guidance knowledge in the paper since it contains both transferred  and noisy clues.

Once we obtain the soft targets, the noisy subset is supervised by KL-divergence Loss as implemented in most of deep learning toolbox. Formally, it is defined as,
\begin{equation}\label{eq:loss1}
\mathcal{L}_{g}(\theta) =  \sum_{i}^N{g}_{i}log(\frac{{g}_{i}}{{q}_{i}}),
\end{equation}
where ${q}$ denotes the softened prediction of networks, ${g}$ is the soft target label.

%\subsubsection{Multi-task Learning}
Finally, the student network is trained by integrating the KL-divergence loss for the noisy subset and cross entropy loss for the clean subset. The total loss is as follows:
\begin{equation}\label{eq:loss2}
\mathcal{L}_{total}(\theta) = \alpha  T^{2}  \mathcal{L}_{g} + \mathcal{L}_{c}
\end{equation}
where $T^{2} $ is used to compensate the impact of soften operation in Eq.(\ref{eq:equ1}), $\alpha$ is the hyper-parameter which balances the importance between these two tasks, and $\mathcal{L}_{c}$ is the cross-entropy loss on clean dataset whose formulation is the same with Eq.(\ref{eq:CE}).% \llc{I do not understand where $T^2$ is used here. and should have been defined. } 

%   \color{1}{tools}, \color{2}{baby and kids}, \color{3}{home}, \color{4}{beauty and personal care}, \color{5}{shoes}, \color{6}{clothing}, \color{7}{accessories}, \color{8}{sports/outdoors}, \color{9}{grocery}, \color{10}{health}, \color{11}{luggage}, \color{12}{electronics} \color{13}{and other}.}
%\begin{figure*}[tp]
%\centering
%%\centering{
% % Requires \usepackage{graphicx}
% \includegraphics[width=0.95\linewidth]{confusion_matrix5.jpg}\\
% \caption{The confusion matrix of our method on Product-90. (Zoom in for better view.)}\label{fig:cm} %The class indexes are corresponding to the categories in Figure \ref{fig:figure6}.
%\end{figure*}

\begin{table}[tp]
\centering
\caption{Comparison on Products-90.}\label{tab:table1}
\scriptsize
\begin{tabular}{llllc}
\toprule
  Model \# & Method & Training Data & Initialization & Test Accuracy \\
  \midrule
  1 & ResNet-101  & noisy data $D_n$ & ImageNet & 60.97$\%$ \\
  2 & ResNet-101   & clean data $D_c$& ImageNet & 62.15$\%$ \\
  3 & ResNet-101  & $D_n$ and $D_c$ & ImageNet & 66.78 $\%$ \\
  4 & Guidance Learning &$D_n$ and $D_c$ & model\#3 & \textbf{68.86} $\%$ \\
%  \midrule
%  5 & Baseline + $D_c$ Finetuning & $D_c$ & model\#3 & 69.60 $\%$ \\
%  6 & Guidance Learning + $D_c$ Finetuning & $D_c$ & model\#4 & \textbf{71.40} $\%$ \\
\bottomrule
\end{tabular}
\end{table}

\section{Experiments}
In this section, we first present the implementation details, and then conduct extensive evaluations with our guidance learning method on the Products-90, and finally apply our method on Food-101 and Food-101N. %We also compare to the state of the arts on these noisy datasets.

\subsection{Implementation Details}
We implement our method with Pytorch. For data augmentation, we resize images to scale 256$\times$256, and randomly crop regions of 224$\times$224 with random flipping. We crop the middle 224$\times$224 regions for testing.
We use ResNet-101~\cite{he2016deep} architecture on Products-90, and ResNet-50 on Food-101N and Clothing1M. All networks are pre-trained on the ImageNet dataset. %
In the teacher network training step, we initialize the learning rate ($lr$) to $10^{-3}$, and divide it by 10 after 10, 15 and 20 epochs. We stop training after 25 epochs. %
In the student network training step, we set the $lr$ to $10^{-4}$, and divide it by 10 after 5, 8 and stop training after 11 epochs.
We use the SGD method for optimization with a momentum of 0.9 and a weight decay of $10^{-3}$. The batch size is set to 64 for all steps.
For the hyper-parameters $\alpha, \beta,$ and T in our guidance learning framework, the default values are 0.1, 0.3, and 5, respectively.

\subsection{Exploration of Guidance Learning on Products-90}
In this section, we first compare our guidance learning method to several well-known baseline methods, and then evaluate the hyper-parameters.

Table \ref{tab:table1} presents the test accuracy comparison between our methods and several baselines, i.e. model \#1, \#2, and \#3. These baselines ignore the noisy label problem and view all labels as ground truth. Traditional cross-entropy loss are used for all these baselines. 
As shown in Table \ref{tab:table1}, training on the noisy set $D_n$ gets the worst result 60.97\%, which even inferior to the one trained on the small clean set (i.e., 62.15\%). It suggests that i) the noisy subset may show different distribution compared to the clean one and ii) noisy labels degrade CNNs significantly even there is a large scale of data. 
As found in \cite{xiao2015learning}, training on both clean and noisy sets is a better choice which achieves 66.78\% on our collected dataset. We use this model as the teacher model of our guidance learning framework.
With the same full dataset, our guidance learning framework further improves the teacher model by 2.08\%.
As another useful trick in noisy data learning~\cite{patrini2017making,guo2018curriculumnet}, fine-tuning the model trained with noisy data on clean set further boosts the final performance.This trick improves our guidance learning from 68.86\% to 71.4\%, and boosts the model\#3 from 66.78\% to 68.6\% .
%\pxj{To further analyze the per-class performance of our dataset, we illustrate the confusion matrix in Figure \ref{fig:cm}, which is generated by our guidance learning method (i.e. the 4-th model in Table \ref{tab:datacomp1}). Several observations can be concluded as follows. First, the first three categories which refer to the ``tools'' meta category are well-classified by our model. It can be mainly explained that these classes like packing box and accessory are very different from other products. Second, the ``clothing'' meta category is the most difficult one. For example, pajamas, dresses, women tops, and dress suits are confused each other.}

%

\begin{figure}[t]
\centering
%\centering{
 \includegraphics[width=0.9\linewidth]{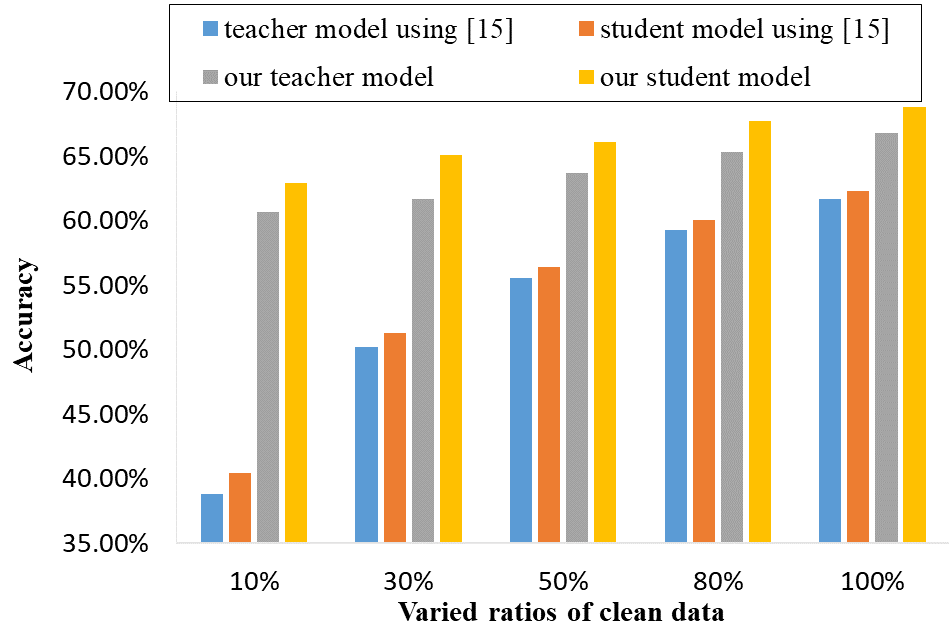}\\
 \caption{Evaluation of clean image ratios w.r.t. the original clean set.}\label{fig:figure7}
\end{figure}

\textbf{The effect of noise}. To investigate the effect of noise for our guidance learning framework, we change the ratio of clean images. Specifically, we reduce the number of clean images to 10\%, 30\%, 50\%, and 80\% of the original clean training set (i.e. $D_c$). Figure \ref{fig:figure7} presents the results of the teacher models and student models on the test set. We also compare our method to the most related work in \cite{li2017learning} which also uses knowledge distillation where the teacher model is trained on clean data and the student model on the full data with modified soft labels. We do not use knowlegde graph to refine soft labels as \cite{li2017learning} since we do not have.
 Our teacher model is slightly impacted by the decreasing of clean images (i.e. we remove partial clean training images). Our guidance learning framework consistently improves the teacher model by more than 2\%. \cite{li2017learning} is inferior to our method consistently and degrades significantly when the ratio of clean images is reduced. For example, reducing the number of clean images to 10\% (about 10 images for each class) degrades about 20\%, and both the teacher model and student model achieve less than 40\%. However, it only leads to 2.22\% degradation (66.64\% vs. 68.86\%) for our student model which demonstrates the robustness and efficiency of guidance learning framework even with tiny-scale of clean images. This can be explained by that i) training with tiny-scale clean data, leads to overfitting easily and ii) a bad teacher model impacts the final performance of its student models.

%Student model\_1 is based on the different teacher models and student model\_2 is based on the 100\% teacher models. 
%With the student model\_1, our method leads to more than 2\% improvements compared with the teacher models. With the student model\_2, reducing the number of clean images to 10\% (about 10 images for each class) only leads to 2.22\% degradation (66.64\% vs. 68.86\%) for our method which demonstrates the efficiency of guidance learning framework even with tiny-scale of clean images.

%The validates that  %

\begin{figure}[t]
\centering
%\centering{
 \includegraphics[width=0.9\linewidth]{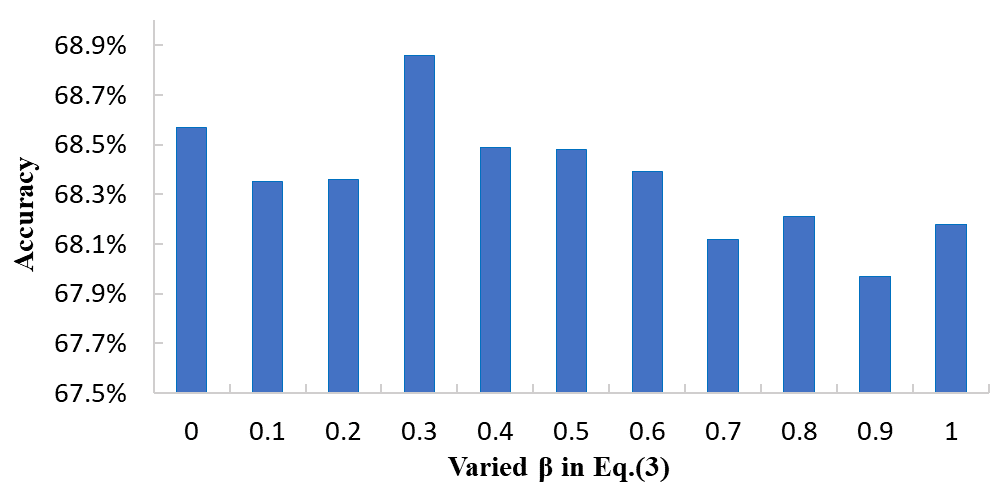}\\
 \caption{Evaluation of $\beta$ in Eq.(\ref{eq:equ2}).}\label{fig:figure4}
\end{figure}

\begin{figure}[t]
\centering
%\centering{
 \includegraphics[width=0.9\linewidth]{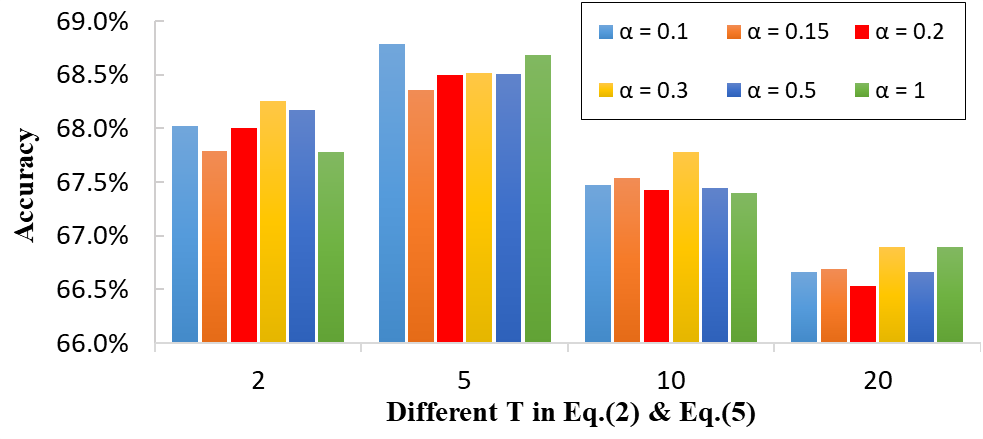}\\
 \caption{Evaluation of $\alpha$ and T in Eq.(\ref{eq:loss2}).}\label{fig:figure5}
\end{figure}

\textbf{Evaluation of $\alpha, T$ and $\beta$}.
 $\beta$ is a trade-off weight between noisy labels and the predictions of teacher model in our guidance learning. Taking the default values of $\alpha$ and T, we evaluate different $\beta$ from 0 to 1. The results are shown in Figure \ref{fig:figure4}. We observe that increasing $\beta$ boosts performance but saturates above 0.3. $\alpha$ balances the importance between the losses of noisy set and clean set, T is the temperature used for softening.
 We evaluate $\alpha$ and T by fixing $\beta$ to 0.3. The results are illustrated in Figure \ref{fig:figure5}. From Figure \ref{fig:figure5}, several observations can be found. First, `T=5' consistently outperforms the others regardless of $\alpha$. Second, increasing $\beta$ boosts performance in the beginning but degrades after 5, which indicates that a highly-soften operation corrupts supervision knowledge.
Third, $\alpha$ and T impact performance jointly which change the loss of noisy set in Eq. (\ref{eq:loss2}). %We achieve the best result with $\alpha=0.1$ and T=5.

\begin{table}[htp]
\centering
\caption{Comparison between our method and recent state-of-the-arts on Food-101. $^*$60K extra images have been used which have both correct labels and noisy labels introduced in \cite{lee2017cleannet}. It is worth noting that 60K is comparable to the number of training images in Food-101. $^\dagger$BNInception is used for backbone network which is pre-trained on the full ImageNet dataset with 21k classes. \textit{The results in brackets are our best reimplementation of CleanNet using Pytorch.}
}\label{tab:table2}
\scriptsize
\begin{tabular}{p{10pt}p{80pt}p{40pt}p{20pt}p{40pt}}
\toprule
  \# & Method & Training Data & Init. &  Accuracy  \\
\midrule
  1 & ResNet-50\cite{lee2017cleannet} & Food-101N & ImageNet & 81.44  \\\midrule
  2 & ResNet-50 \cite{lee2017cleannet} & Food-101 & ImageNet & 81.67  \\\midrule
  3 & ResNet-50 & Food-101 and Food-101N & ImageNet & 85.80  \\\midrule
  4 & ResNet-50 & Food-101N$^*$ & ImageNet & 79.83  \\\midrule
  5 & CleanNet(hard)~\cite{lee2017cleannet} & Food-101N$^*$ & ImageNet & 83.47(\textit{82.39})  \\\midrule
  6 & CleanNet(soft)~\cite{lee2017cleannet} & Food-101N$^*$ & ImageNet & 83.95(\textit{82.99})  \\\midrule
  7 & Curriculum~\cite{guo2018curriculumnet}$\dagger$ & Food-101 and synthetic data  & ImageNet & 87.3  \\\midrule
  8 & Guidance Learning & Food-101N$^*$ & model\#4 & 84.20  \\\midrule
  9 & Guidance Learning & Food-101 and Food-101N & model\#3 & \textbf{87.36}  \\
%  10 & Guidance Learning + $D_c$ Finetuning & Food-101 & model\#9 & \textbf{87.88}  \\
%    & Baseline/Teacher + $D_c$ Finetuning & Food-101 & & model\#3 & 86.69  \\
\bottomrule
\end{tabular}
\end{table}

\begin{table*}[!t]
\centering
\caption{Performance comparison between our method and recent state-of-the-art methods on  Clothing1M. $^\ast$32K images have both correct labels and noisy labels which are used to train CleanNet. $^\ddagger$BNInception is used for backbone network which is pre-trained on the full ImageNet dataset with 21k classes.}\label{tab:table3}
%\scriptsize
\begin{tabular}{llllc}
\toprule
  \# & Method & Training Data & Initialization &  Accuracy\\
\midrule
  1 & ResNet-50 & 1M noisy  & ImageNet & 68.94  \\
  2 & ResNet-50  & 50K clean & ImageNet & 75.19  \\
  3 & ResNet-50  & 1M noisy + 50K clean & ImageNet & 71.61  \\
  4 & CleanNet(hard) \cite{lee2017cleannet} & 1M noisy$^\ast$& ImageNet & 74.14  \\
  5 & CleanNet(soft) \cite{lee2017cleannet} & 1M noisy$^\ast$ & ImageNet & 74.69  \\
  6 & CleanNet(soft)~\cite{lee2017cleannet}+ $D_c$ Finetuning & 50k clean & model\#5 & 79.90  \\
  7 & Loss correction~\cite{patrini2017making} & 1M noisy & ImageNet & 69.84  \\
  8 & Loss correction~\cite{patrini2017making} + $D_c$ Finetuning& 50k clean & model\#7 & 80.38  \\
  9 & Curriculum \cite{guo2018curriculumnet}$^\ddagger$  & 1M noisy  & ImageNet & 75.80  \\
  10 & Curriculum \cite{guo2018curriculumnet}$^\ddagger$  + $D_c$ Finetuning & 50k clean  & model\#9 & \textbf{81.50}  \\
  11 & Guidance Learning & 1M noisy + 50K clean & ImageNet & 75.76  \\
  12 & Guidance Learning + $D_c$ Finetuning & 50k clean & model\#11 & 80.31  \\
  13 & Guidance Learning $^\ddagger$ & 1M noisy + 50K clean & ImageNet & 78.77  \\
  14 & Guidance Learning + $D_c$ Finetuning $^\ddagger$& 50k clean & model\#13 & \textbf{81.13}  \\
  \bottomrule
\end{tabular}
\end{table*}

\subsection{Experiments on Food-101 and Food-101N}
\textbf{Food-101 and Food-101N}. The Food-101 dataset~\cite{bossard2014food} is a benchmark for visual food evaluation. It contains 101 food categories, with 101,000 real-world food images totally. For each class, 750 images are used for training, the other 250 images for testing. It is a clean dataset reviewed manually. The training set of Food-101 is used as clean trainset $D_c$. To conduct experiments with label noise, we utilize the Food-101N noisy dataset as $D_n$.  The Food-101N  dataset~\cite{lee2017cleannet} is collected from Google, Bing, Yelp, and TripAdvisor with the concepts of Food-101, and is filtered out from foodspotting.com where the Food-101 is collected. We use all the 310k images of Food-101N as the noisy dataset in our experiments, and report the overall accuracy on the Food-101 test set.

We futher validate our method on Food-101 and Food-101N. The performance comparison is presented in Table \ref{tab:table2}. The first two baselines are provided in \cite{lee2017cleannet}. Our teacher model trained on both the training set of Food-101 and Food-101N obtains 85.8\% which outperforms the baselines and CleanNet~\cite{lee2017cleannet} in a large margin. It demonstrates that more data is better even there has label noise. Our guidance learning method improves the baseline result from 85.8\% to 87.36\%, which outperforms the current state-of-the-art methods.
It is worth noting that CleanNet, which is a state-of-the-art semi-supervised noisy data learning method, leverages additional 60K manually-verified images within the Food-101N. The number of 60K is comparable to 75K of the original training images on Food-101. For a fair comparison, we conduct an extra experiment with the same 60K manually-verified set, and obtain 84.2\% which is better than both the hard (82.39\%) and soft (82.99\%) version of CleanNet. 
%We achieve 87.88\% by finetuning the guidance learned model on Food-101, which sets the new state of the art.

%We now examine the performance of the proposed approaches on Food 101. As the described of the Food 101 in $Section Public data$, we use the clean training and testing data from the Food101 and the noisy training data from the Food-101N. Therefore, we follow the setting and compare with the baseline results on \cite{lee2017cleannet}.
%
%Table \ref{tab:table2} illustrated the results of our experiments. We use all training data to obtain the teacher model can show about $+4\%$ improvement in accuracy compared to the results of the baseline in row \#1 and \#2. It is also better than all the results on the CleanNet Method, which is shown in row \#4 and \#5. Additionally, it can bring approximately $3\%$ enhancement than the best results on CleanNet Method. Based on the model \#3, our method can achieve $87.36\%$ and outperform the  baseline of model\#3 and the Curriculum method \cite{guo2018curriculumnet} in row \#6. We also use the clean training data from the Food101 to finetune our proposed model in row \#7. The final result can up to $87.88\%$, which is the state of art classification performance on the Food101.

%
\subsection{Experiments on Clothing1M}
\textbf{Clothing1M}. The Clothing1M dataset~\cite{xiao2015learning} is a public large-scale fashion dataset to evaluate recognition accuracy from noisy data with human supervision. It contains 1 million images with noisy class labels from 14 fashion classes and thousands of human-annotated images. All the images are crawled from several online shopping websites. The human-annotated set is used as the clean set which is further split into training $D_c$, validation and test sets with the size of 50k, 14k, 10k, respectively. A confusion matrix between the human annotations and the original noisy labels shows that the overall accuracy is 61.54\% in~\cite{xiao2015learning}. We report the overall accuracy on the test set of Clothing1M.

%We evaluate our method on the widely-used Clothing1M dataset. The performance comparison is shown in Table \ref{tab:table3}.
The first two baselines are provided in \cite{patrini2017making}. Our baseline trained on mixed noisy and clean data is 71.61\% which is slightly lower than those in \cite{patrini2017making}. Both \cite{lee2017cleannet} and \cite{patrini2017making} use ResNet-50 as backbone network, and respectively obtain 74.69\% and 69.84\% without the final finetuning process on clean data. With the same backbone, our guidance learning method improves the teacher model from 71.61\% to 75.76\% which outperforms \cite{lee2017cleannet} and \cite{patrini2017making}.
CurriculumNet~\cite{guo2018curriculumnet} uses the BNInception~\cite{ioffe2015batch} network which is pre-trained on the full ImageNet with 21k classes as its backbone model, and is implemented with Caffe\footnote{http://caffe.berkeleyvision.org/}. For a fair comparison, we replace the ResNet-50 as the same BNInception model, and obtain 78.77\% without finetuning on clean set which is 2.93\% better than CurriculumNet (78.77\% vs. 75.8\%).

For the results with a further finetuning on clean set, the accuracies are 79.9\%, 80.38\%, and 81.5\% in \cite{lee2017cleannet}, \cite{patrini2017making}, and \cite{guo2018curriculumnet}, respectively. We obtain 81.13\% with the finetuning trick which is comparable to the state of the arts. %Moreover, with the same backbone of CurriculumNet, we can get better result which is up to 81.13\%.
\cite{patrini2017making} estimates a confusion matrix which indicates the probability of each class being corrupted into
another. This method is based on the assumption that the noisy label can be corrected to another. \cite{guo2018curriculumnet} designs a curriculum to train noisy data from easy to hard which utilizes a clustering process and a training process repeatedly. Compared to these two state-of-the-art methods, our guidance learning method is more efficient and simpler.

\section{Conclusion}
In this paper, we present a large-scale daily product image dataset, termed as Product-90, for recognizing product in daily life. Compared to exsiting product datasets, our product dataset is more diverse in product categories and owns more images. Since our Product-90 introduces noisy labels, we also propose a simple yet efficient guidance learning framework to address the problem of training CNNs from noisy data.
It first trains an initial teacher network from the full dataset including both clean and noisy data, and then separates the noisy part and clean part, and finally trains a target network with multi-task learning. In the target network training step, the noisy data is supervised by the guidance knowledge which is the combination of its noisy label and soft label from the teacher network. Experiments on several public datasets and our dataset show that our guidance learning method improves the base model significantly and achieves state-of-the-art performance. All the code will be publicly available including the reimplementation of CleanNet.
%In addition, we introduce a real-world noisy dataset named Product-90 which is collected from the image-text customer reviews of several online e-shopping websites. Compared to existing noisy datasets, our Product-90 is more challenging due to severe background clutter, visually confused images, fine-grain issues and so on.

{\small
\bibliographystyle{ieee}
\bibliography{article}
}

\end{document}

%% file: intro2.tex
\section{Introduction}

%In recent years, smart retail is becoming a dramatically growing market. It is reported that the market size of smart retail is 13.07 billion USD in 2018, and will  reach USD 38.51 billion by 2023 \footnote{https://www.researchandmarkets.com/reports/4602326/smart-retail-market-by-application-visual}.
%The development of smart retail industry provides both opportunities and challenges for computer vision research community.

This paper studies a crucial problem in real-world application: recognize products from consumer photos without much supervision. More specifically, we want to recognize the fine-grained products taken by consumer's mobile cameras, with unconstrained viewing directions, cluttered background, and different lighting conditions. One can imagine an application that you are recommended where the products can be found and what the prices are by recognizing your casually-captured product photos. 
%In practice, the product list is keep changing and consumer photos are updated very quickly. It is often impossible to label enough samples for training, and hence presents the challenge of learning with limited supervision. 
\begin{figure}[tb]
\centering
%\centering{
  \includegraphics[width=\linewidth, height=0.8\linewidth]{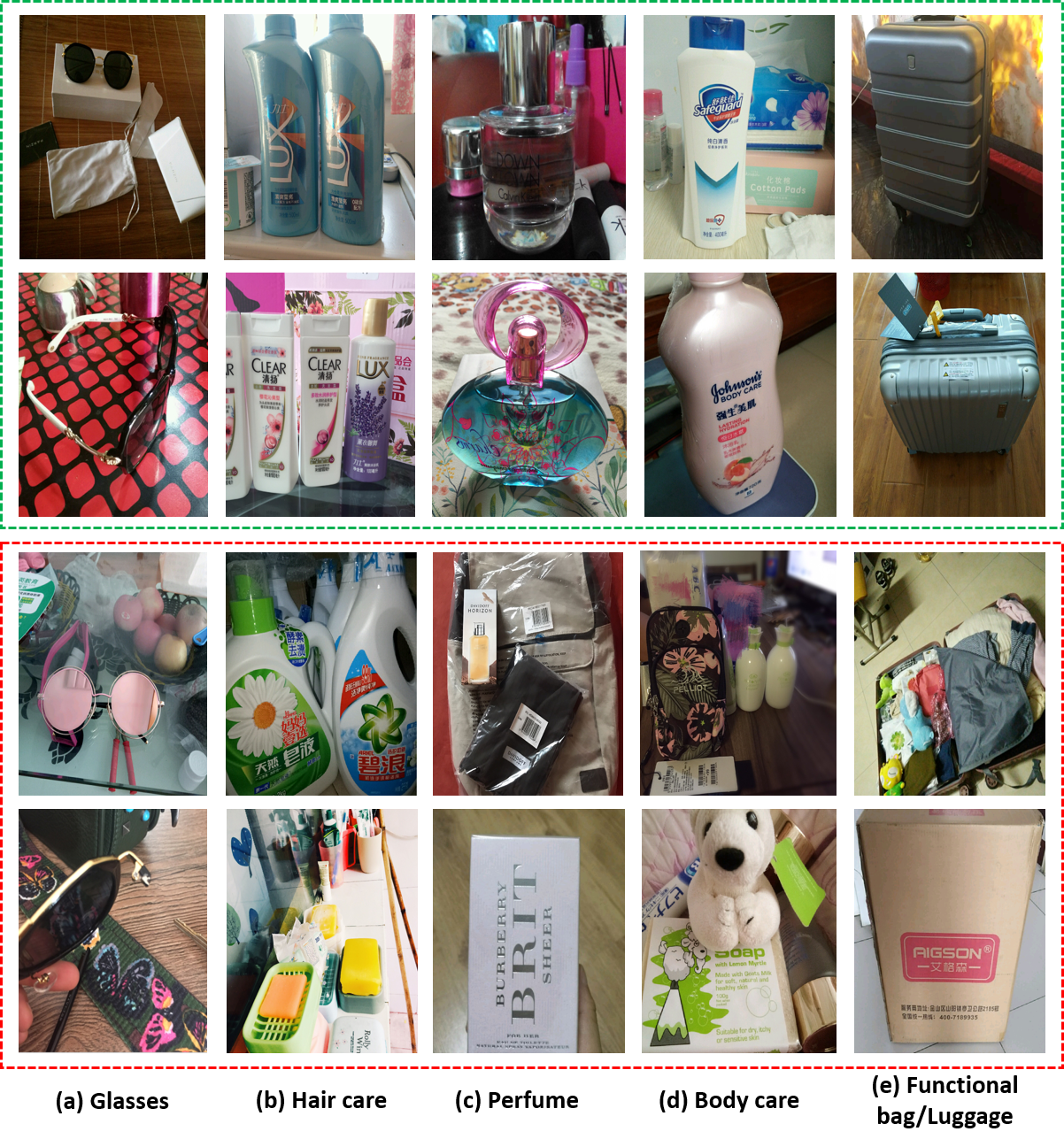}\\
  \caption{Example images from our Products-90. We illustrate 5 different categories in column. Visually correct images are shown in the first two rows. Visually confused or unrelated images are shown in the last two rows.}
\label{fig:figure1}
\end{figure}
% \llc{this paper does not look clear or beautiful enough. The reader may get lost since it tells a lot of things together: examples, noise, and etc. To make it clear, we may separate it into whole-page width or separate it into two figures.}  \llc{shall we change it to unrelated coz the latter is more popular?}

To address this real-world product image recognition task,
we build a novel large-scale dataset, termed as \textsl{Product-90}, which consists of 90 generic product categories. Instead of collecting daily images by labor-and time-intensive image capturing, we take advantage of the web and download images from the reviews of several e-commerce websites where the images are casually captured by consumers.
Totally, we collected
more than 140k product images from the customer reviews. 
The associated 90 categories are borrowed the categories of e-commerce websites. Figure \ref{fig:figure1} shows some examples of this dataset. We can see there are several challenges brought by \textsl{Product-90} dataset:
i)The visual contents in Product-90 contains a wide range of subjects. ii) Some categories are very similar in appearance, e.g. Hair Care vs. Body Care. iii) Some photos are not related to the category, which suggests a significant level of noise exists in the dataset. 

To evaluate product image recognition algorithms on our Product-90, we build a small manually-clean subset for traditional training and testing, and remain the rest of Product-90 as noisy data which can be used for extra training.  To take full advantage of the small clean training subset and the massive noisy labeled data for daily product recognition, we propose a novel \textit{guidance learning framework} for noisy data learning. It mainly includes two training stages.
At the first stage, we train a baseline CNN model, or a teacher model, on the full Product-90 dataset (without the clean test set).
At the second stage, we train a student or target network on the large-scale noisy set and the small clean training set with multi-task learning. Specifically, in the stage of student training, the large-scale noisy data is supervised by the guidance knowledge which consists of two supervision signals, namely the noisy ground truths and the soften labels from the teacher network. We fuse these one-hot ground truths with the soften multi-hot labels, and optimize the network by Kullback--Leibler Divergence (KLDiv) loss. % between the soft and predicted distributions.

Our guidance learning framework considers the following issues.
The first stage of our guidance learning ensures that we can obtain a powerful teacher model instead of using the noisy set or clean set only like in \cite{patrini2017making}.% which is vital to obtain reliable soft labels for training in the next stage.
We fuse both ground truths and soften labels for the large-scale noisy data, since i) the teacher model provides useful information but is far from perfect, and ii) there exist both false and correct labels in noisy labels.

In summary, our contributions can be concluded as follows: 
\begin{itemize}
    \item We introduce a new task, i.e. daily product image recognition, and a novel large-scale dataset, termed as Product-90 which is collected from the reviews of e-commerce websites.
    \item To advance the performance of daily product image recognition, we propose a generic guidance learning method to take full advantage the small clean subset and the large-scale noisy data in Product-90.
    \item We conduct comprehensive evaluations with our guidance learning method on our Products-90, Food-101~\cite{bossard2014food},  Food-101N~\cite{lee2017cleannet}, and Clothing1M~\cite{xiao2015learning}, and achieve state-of-the-art results. 
\end{itemize}